\pdfoutput=1
\documentclass[conference]{IEEEtran}
\IEEEoverridecommandlockouts
\usepackage{cite}
\usepackage{amsmath}
\usepackage{amsmath,amssymb,amsfonts}
\usepackage{algorithmic}
\usepackage[linesnumbered,ruled,vlined]{algorithm2e}
\usepackage{graphicx}
\usepackage{textcomp}
\usepackage{xcolor}
\usepackage{multicol}
\usepackage{multirow}
\usepackage{hyperref}
\hypersetup{colorlinks,allcolors=black}
\usepackage{tabularx}
\def\BibTeX{{\rm B\kern-.05em{\sc i\kern-.025em b}\kern-.08em
    T\kern-.1667em\lower.7ex\hbox{E}\kern-.125emX}}
\begin{document}

\title{Stacked Generalization for Human Activity Recognition\\}

\author{\IEEEauthorblockN{Ambareesh Ravi}
\IEEEauthorblockA{
\textit{Center for Pattern Analysis and Machine Learning}\\
\textit{Dept. of Electrical and Computer Engineering} \\
\textit{University of Waterloo, Waterloo, CANADA}\\
ambareesh.ravi@uwaterloo.ca}}

\maketitle

\begin{abstract}
This short paper aims to discuss the effectiveness and performance of classical machine learning
approaches for Human Activity Recognition (HAR). It
proposes two important models - Extra Trees and Stacked
Classifier with the emphasize on the best practices,
heuristics and measures that are required to maximize
the performance of those models.\\
\end{abstract}


\section{Introduction}
This work primarily focuses on the usage of
lesser known, unconventional algorithms that can
significantly enhance performance on complex
data. Many classical machine learning algorithms
suffer from the problem of overfitting. Sometimes
the train data may not be a good representation of
real time data but many times the model could be
insufficient in terms of capacity to learn complex
patterns in data. The ensemble methods proposed
in this work, not only increase the accuracy of prediction but also the reliability of the performance
on unseen data.

\section{DATA SET AND FEATURE ENGINEERING}
The data set taken for this study is UCI’s HAR
data set \cite{1} which consists of human activity data
recorded from a smartphone that spans across 6
classes. Each sample consists of 561 features from
sensors in a smartphone that are processed and
filtered. The data is split into two parts randomly -
70\% for training and 30
Many of the features were found to be co-dependent and reducing the number of features
could maximize the performance of the model
by exposing it to the important features to learn.
Feature (dimensionality) reduction was done using
Principal Component Analysis (PCA), a simple
and reliable algorithm. \textit{The number of components
for PCA was selected as 200 for a Proportion of
Variance of 99.75\% to capture maximum information and the data was reduced.}

\section{MODEL SELECTION}
For the task of classification, there is a palette
of classical machine learning algorithms to chose
from and the most important ones are reviewed
and listed with mean accuracy in this section in Table \ref{tab: Table A}.

\begin{table}
    \centering
    \begin{tabular}{|c|c|c|c|}
        \hline
        Model / PCA & 200 & 400 & False\\
        \hline
        Bagging & 0.883 & 0.873 & 0.897\\
        Decision Tree & 0.799 & 0.781 & 0.862\\
        Extra Trees & 0.919 & 0.880 & 0.942\\
        Gradient Boosting & 0.926 &  0.927 & 0.939\\
        K Nearest Neighbors & 0.903 & 0.901 & 0.901\\
        Log. Reg. OneVSRest & 0.962 & 0.961 & 0.962\\
        Random Forest & 0.911 & 0.897 & 0.930\\
        SVM-Linear & 0.965 & 0.966 & 0.965\\
        SVM-Rbf & 0.947 & 0.946 & 0.940\\
        \hline
    \end{tabular}
    \caption{\label{tab: Table A} Effect of PCA on various models}
\end{table}

\textbf{The models with the best individual performance were shortlisted for the next stage of the
work}. A review of important methods are given
below:

\subsection{Logistic Regression}
Logistic Regression \cite{5} is a statistical method that estimates the probability
of occurrence of one more categorical variable
with respect to others using a logistic function
which is defined by the sigmoid function in equation \ref{sigmoid}.

\begin{equation}
    f(x) = \frac{1}{1+e^{-x}}
    \label{sigmoid}
\end{equation}
The logistic function directly calculates the prob-
ability of an occurrence as the value of function
ranges between 0 and 1. The logistic regression
uses a hypothesis function to learn and calculate
the probabilities whose parameters can be estimated using a loss function and an optimizer. The
hypothesis, loss and optimizing functions can be
chosen based on the problem at hand.

\subsection{Gradient Boosting}
Gradient Boosting \cite{6} is
an ensemble technique that typically uses Decision
Trees (usually regression trees). Boosting is the
process of filtering out data samples that are learnt
to focus more on the ones that are not. Gradient
Boosting adds weak learners one by one in an additive nature and an optimizer is used to minimize the loss while adding trees using a loss function
to check the correctness of the predictions of the
subsequent trees. The contribution from each tree
is weighted or shrunk using the optimizer while
learning. Both loss and optimizer can be chosen
based on the data and problem setup.

\subsection{State Vector Machines}
State Vector Machine\cite{4} generally works to find a linear separation in data and in the absence of which it projects data to higher dimensional space to separate the data linearly. It works by maintaining and maximizing
a margin of separation between the data which
is supported by closest points of different classes
called ’support vectors’ . Depending on the nature
of the margin to allow some overlap of samples
/ misclassifications, SVM is categorized into soft
and hard margin SVMs. SVM focuses mostly on
differentiating the data points that are the very similar in nature in comparison to the other classifiers
that pay attention to all of the points.

\section{REVIEW AND MODEL DESCRIPTION}
Each of the models in the above list were
found to perform well on the data. To improve the
performance further, this project deals with two important ensemble methods Extra Trees Classifier
and \textbf{Stacked Generalization Classifier}.\\

Ensemble methods use multiple ’weak’ learners
or models to learn parts or whole of the data
and combine the results from individual models
to produce the final prediction to achieve the
best possible performance. Ensemble methods are
proven to have the following advantages \cite{7}[**1]:
\begin{enumerate}
    \item Combining predictions from multiple models
in ensembles works better since the predictions from the sub-models are at best weakly
correlated as each model learns a subset of
data and not the whole data.
    \item Variance and/or Bias can be reduced.
    \item Greedy or Local learning can be compensated
for, with other algorithms to improve generalization.
    \item A variety of different algorithms can be chosen and used in parallel for learning and
prediction.
\end{enumerate}

\subsection{Extra Trees Classifier}
Extra Trees Classifier or \textbf{Extremely randomized classifier} is a tree based ensemble method which uses Decision Trees as the primary components with top-down approach. It is similar to Random Forest in many aspects but the two
key differences are that \textit{features to split on are
chosen at random at all stages and there is no
bootstrapping used while drawing samples}. It is
very useful while operating with a large number
of varying continuous features since it reduces the
burden of computing the best feature to split on
by choosing at random. The bias/variance analysis
has shown that Extra-Trees work by decreasing
variance while at the same time increasing bias
\cite{2}. Some of the salient features of Extra Trees
Classifier are given below[**2]:
\begin{enumerate}
    \item Computationally efficient owing to the simplicity of node splitting procedure.
    \item Without bootstrapping, we reduce the repetition of data in the learning process.
    \item Works on both categorical and numerical variables with the ability to have multiple random splits at each node depending on the data and problem setup.
    \item Even though randomization increases the variance of individual weak learners, it decreases the overall variance of the ensemble with respect to each sample. Also, the randomization can help the model generalize and avoid overfitting.
\end{enumerate}
Though the nature of operations of Extra Trees
Classifier is close to that of a Random Forest Classifier, there are significant performance differences
between the two - Refer to the \textit{results section.}

\subsection{Stacked Generalization Classifier}
The \textbf{Stacked (Generalization) Classifier} is another ensemble method that is used to combine the
best properties of multiple models into a single
best entity. In Stacked Classifier, the information
learnt is passed from one level to another level
to predict the final desired outcome \cite{3}. Multiple
classifiers are learned together and a final decision
is made based on the output of all the classifiers
in the stack by another classifier (Usually Logistic
Regression) which is called the \textit{meta-learner}.

Stacked Generalization by nature, works by reducing the bias of the individual models used in the stack \cite{3}, hence improves the generalization of the model and reduces the error rate significantly. It can also be thought of a replacement for cross
validation where the best model on the validation
data is chosen which might not be the overall best
model \cite{3}. But here, all the models are made to
learn on parts of data and their overall performance
is improved.[**3].

The parameters of individual models directly
impact their performance on any given data.
For example, number of neighbors (K) in KNN,
number of trees, max depth in tree based
methods, learning rate and optimizer in gradient
based methods, regularization parameter (C) in
SVM etc. and other factors like the first feature
to split, number of tune-able weights in Logistic
Regression etc. All these factors create strong
bias towards the predictions of the model with
respect to unseen data. All these are averted by
using multiple classifier that perform well on the
parts of data they learnt individually which is in
the nature of stacked classifier.[**4]

\begin{algorithm}
\SetAlgoLined
Define ’\textbf{T}’ classifiers for the stack.\\

Learn T classifiers on data set ’\textbf{D1}’.\\

Learn another level of classifier ’\textbf{g}’ from data set ’\textbf{D2}’ using meta-data ’\textbf{M}’ defined by $(f_t(x_i ), y_i)$ where $f_t(x_i)$ is the output of $t^{th}$ classifier for $i^{th}$ sample of data, $\forall t \epsilon T$.\\

Predict the output of new classifier g where $y_i = g(f_t (x_i))$ for t = 1..T
\caption{Algorithm for Stacked Ensemble}
\end{algorithm}

\subsection{Best Practices}
Theoretically, any number of levels can be added but it is wise to chose the
number of levels depending on the performance
of each level. Two levels are found to be sufficient
to solve most of the problems \cite{3}. In practice,
domain knowledge on the data will be helpful
in configuring the model based on the nature of
data and capacity of individual models. Careful
choice of individual models and their respective
configurations can significantly help in improving
the performance over any other individual models.

Tuning the stacked classifier can prove to be
difficult owing to the number of combinations
with respect to the hyperparameters from all the
individual models in the stack. So it is heuristic to tune and check the individual learners first and to
proceed implementing the stack.

\subsection{Methodology}
So, after carefully considering
the performance of the models tested previously
and the stipulation to chose a novel, unconventional model, Stacked Classifier was chosen with
the following individual \textbf{models and best parameters} based on individual performance based on 10 times 10 fold cross validation in Table \ref{tab:best_params}

\begin{table}[]
    \centering
    \begin{tabular}{|c|c|}
        \hline
        Model & Best Parameters \\
        \hline
        Logistic Regression & OneVSRest, l1 norm\\
        Linear SVM & C = 2.0\\
        Gradient Boosting & n estimators = 50, Learning
rate = 0.2\\
        Extra Trees Classifier & n estimators = 100, max
depth = 4\\
        \hline
    \end{tabular}
    \caption{Models and best parameters}
    \label{tab:best_params}
\end{table}
While fitting or training the data, the above four
models in the stack learn from the data samples
and output their respective predictions. These out-
put predictions of those models which are called
the meta-features will be learned by another classi-
fier (usually and in this case, Logistic Regression)
which is called the meta-learner. It is strongly
advisable to split the data into two parts - one for
training the first layer of the classifiers and other
part to train the meta learner. \textbf{Justifications are
marked with ** throughout the paper.}

\section{Results}
This section discusses the results and findings
of the experiments conducted with Extra Trees and
Stacked Classifiers.

\subsection{Extra Trees VS Random Forest}
Extra trees classifier performed better on the
given data set with a training time which is a
fraction of that of Random Forest’s details of
which are given in the results section.

Performance differences between Random Forest
and Extra Trees classifiers with 200 estimators on
data after 10 times 10 fold cross validation on the
data is shown in Table \ref{tab:rf_et}

\begin{table}[]
    \centering
    \begin{tabular}{|c|c|c|}
        \hline
        Parameter & Random
Forest & Extra
Trees\\
         \hline
        Test Accuracy with PCA & 0.8954 & 0.9090\\
        Training time (seconds) & 20.4 & 4.12\\
        Variance & 0.0012 & 0.0010\\
         \hline
        Test Accuracy without PCA & 0.9294 & 0.9391 \\
        Testing time (seconds) & 22.8 & 4.4 \\
        Variance & 0.0016 & 0.0014 \\
         \hline
    \end{tabular}
    \caption{Performance differences between Random Forest
and Extra Trees classifiers}
    \label{tab:rf_et}
\end{table}

\subsection{Performance of Stacked Classifier}
In terms of overall performance, the stacked
classifier performs better than all the individual
models on the test set except for Linear SVM
which scored 0.06
and the stacked model showed the lowest variance
with respect to all other individual models[**5].
The mean accuracy and variance of the stacked
model after 10 times 10 fold cross validation are
given in Table \ref{tab:acc_var}

\begin{table}[]
    \centering
    \begin{tabular}{|c|c|}
        \hline
        CV Mean Accuracy on Training data & 0.9247 \\
         \hline
        CV Accuracy - Variance Training data & 0.0007\\
         \hline
        Overall Test Accuracy & 0.9562\\
         \hline
    \end{tabular}
    \caption{Mean accuracy and variance}
    \label{tab:acc_var}
\end{table}

\begin{figure}
    \centering
    \includegraphics[width = 0.5\textwidth]{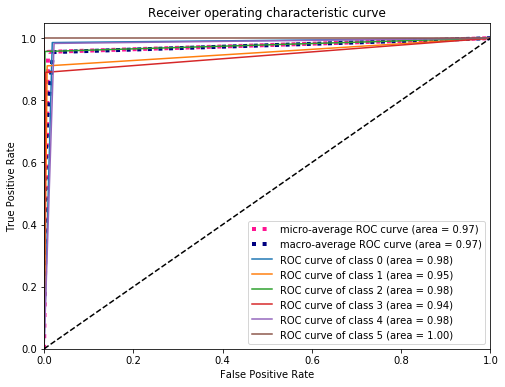}
    \caption{Multi-Class Performance}
    \label{fig:perfr}
\end{figure}

The Multi-class Receiver Operator Characteristics
(ROC) curve and Area Under ROC score, showing
the classification performance of the stacked model
on the test data are shown in figure \ref{fig:perfr}

The
multi-class
confusion
matrix
for the classes in the order walking, walking upstairs,walking downstairs, sitting,
standing, laying of the stacked model on the test
data is shown below:\\

\begin{center}
    $\begin{bmatrix}
    489 & 5 & 2 & 0 & 0 & 0\\
    39 & 429 & 3 & 0 & 0 & 0\\
    6 & 12 & 402 & 0 & 0 & 0\\
    0 & 0 & 0 & 437 & 54 & 0\\
    0 & 0 & 0 & 9 & 523 & 0\\
    0 & 0 & 0 & 0 & 0 & 537\\
    \end{bmatrix}$
\end{center}    

The complete classification report which denotes the class-wise performance of the stacked model is shown in the table \ref{tab:class_repo}

\begin{table}[]
    \centering
    \begin{tabular}{|c|c|c|c|c|}
    \hline
    Class & Precision & Recall & F1-Score & Support \\
    \hline
    Walking & 0.92 & 0.99 & 0.95 & 496\\
    Walking-Upstairs & 0.96 & 0.96 & 0.96 & 420\\
    Walking-Downstairs & 0.98 & 0.89 & 0.93 & 491\\
    Sitting & 0.98 & 0.89 & 0.93 & 491\\
    Standing & 0.91 & 0.98 & 0.94 & 532\\
    Laying & 1.00 & 1.00 & 1.00 & 537\\
    \hline
    Accuracy & & & 0.96 & 2947\\
    Macro Avg & 0.96 & 0.96 & 0.96 & 2947\\
    Weighted Avg & 0.96 & 0.96 & 0.96 & 2947\\
    \hline
    \end{tabular}
    \caption{Classification Report}
    \label{tab:class_repo}
\end{table}

In terms of computational performance, the
stacked model took about\textbf{ 38 minutes} to fit on a $6^{th}$
Gen Core-i5 computer with 12GB memory where
as the other models took around \textbf{25 minutes}. The
prediction time of the model was \textbf{2.3 seconds}. The
model used 1.2GB of memory for fitting inclusive of 200MB for data.

\textit{\textbf{Stacked classifier performs poor when com-
pared to Linear SVM (by 0.06
bias due to low variance (Bias-Variance Trade
Off). Also, Linear SVM finds a higher dimension
to project the data to find a linear separation.}}

\section{FUTURE SCOPE}
The following can be taken up as future works:
\begin{enumerate}
    \item Finding an optimal way to tune the stack
jointly is essential to get the best out of the
models.
    \item Study of the effect of number levels in the
stack on a more complex data could help
understand the potential of the algorithm.
    \item Paralleling computations inside the stack to
reduce the time could be a significant improvement.
\end{enumerate}

\section{CONCLUSION}
This study deals with two ensemble methods -
Extremely Randomized Trees (ExtraTrees) Classifier and Stacked (Generalization) Classifiers for
best performance in terms of speed and accuracy
respectively with heuristics and best practices to
employ them effectively. It also enunciates the
effectiveness of the models with compelling results
in comparison with the conventional ones that are
used and how the stacking scheme improves generalization while learning to increase the accuracy
of prediction on the given data.


\begin{thebibliography}{00}
\bibitem{1} Davide Anguita, Alessandro Ghio, Luca Oneto, Xavier Parra and Jorge L. Reyes-Ortiz. Human Activity Recognition on Smartphones using a Multiclass  Hardware-Friendly SupportVector Machine. International Workshop of Ambient Assisted Living (IWAAL 2012). Vitoria-Gasteiz, Spain. Dec 2012.
\bibitem{2} Pierre Geurts, Damien Ernst, Louis Wehenkel, Extremely randomized trees, Springer Science + Business Media, Inc. 2006, Nov 2005.
\bibitem{3} Wolpert, David H. Stacked generalization. Neural networks 5.2 (1992): 241-259.
\bibitem{4} Cortes, Corinna; Vapnik, Vladimir N. (1995). ”Support-vector networks” . Machine Learning. 20 (3): 273297. CiteSeerX 10.1.1.15.9362. doi:10.1007/BF00994018.
\bibitem{5} Cox, David R. (1958). ”The regression analysis of binary sequences (with discussion)”. J Roy Stat Soc B. 20 (2): 215242. JSTOR 2983890.
\bibitem{6} Friedman, J. H. (February 1999). ”Greedy Function Approximation: A Gradient Boosting Machine” (PDF).
\bibitem{7} Optiz, D et. al. ”Popular Ensemble Methods: An Empirical Study”, Journal of Artificial Intelligence Research 11 (1999).
\end{thebibliography}
\end{document}